\documentclass[letterpaper, 10 pt, conference]{ieeeconf} 
\IEEEoverridecommandlockouts
\usepackage{cite}
\usepackage{amsmath,amssymb,amsfonts}
\usepackage{algorithmic}
\usepackage{graphicx}
\usepackage{newtxtext,booktabs}
\usepackage{tikz,xcolor,hyperref}
\usepackage{tikz,xcolor,hyperref}
\definecolor{lime}{HTML}{A6CE39}
\DeclareRobustCommand{\orcidicon}{%
 \begin{tikzpicture}
 \draw[lime, fill=lime] (0,0) 
 circle [radius=0.16] 
 node[white] {{\fontfamily{qag}\selectfont \tiny ID}};
 \draw[white, fill=white] (-0.0625,0.095) 
 circle [radius=0.007];
 \end{tikzpicture}
 \hspace{-2mm}
}

\graphicspath{{.\figures}}

\foreach \x in {A, ..., Z}{%
 \expandafter\xdef\csname orcid\x\endcsname{\noexpand\href{https://orcid.org/\csname orcidauthor\x\endcsname}{\noexpand\orcidicon}}
}

\begin{document}

\title{\LARGE \bf TouchDrive: Electronics-Free Tactile Sensing Interface for Assistive Grasping
  } 
\author{
        Jing~Xu$^{1}$\orcidA,
        Xuezhi~Niu$^{2}$\orcidB,
        Didem~G\"urd\"ur~Broo$^{2}$\orcidC,
        and~Klas~Hjort$^{1}$\orcidD        
\thanks{$^{1}$Division of Microsystems Technology, Department of Electrical Engineering; 
$^{2}$Cyber-Physical Systems Lab, Department of Information Technology, 
Uppsala University, Uppsala, Sweden. \emph{Corresponding author: Klas Hjort} (\href{mailto:klas.hjort@angstrom.uu.se}{\texttt{klas.hjort@angstrom.uu.se}}).}%
}

\maketitle

\begin{abstract}
Assistive robotic grasping plays an important role in enabling safe and adaptive manipulation of diverse objects. However, existing systems often rely on electronic sensing and multi-stage processing pipelines, increasing system complexity and reducing accessibility. To address these limitations, we present \textit{TouchDrive}, a cost-effective, electronics-free tactile sensing interface for assistive grasping. TouchDrive directly converts contact forces into pneumatic feedback through valve-mediated switching, integrating sensing, signal generation, and feedback within a single passive mechanical loop. The system can be employed using a pneumatic normally closed valve, a compressed air tank, sensing element, and haptic feedback actuator without electronics. By delivering tactile cues, TouchDrive empowers users to modulate grasp forces, enabling precise and robust delicate manipulation of compliant and fragile objects. The interface has been validated across diverse robotic platforms, consistently demonstrating reliable performance and practical applicability in assistive grasping tasks, such as handling fruits and everyday items (up to 20 objects).

\end{abstract}


\section{Introduction}
Haptic interfaces have been widely studied in robotics to improve human perception and control in assistive manipulation and teleoperation~\cite{okamura2004methods, basdogan2020review, lloyd2024pose}. Experimental and user studies show that haptic feedback enhances performance in contact-rich tasks such as surgical manipulation, teleoperated grasping, assembly, and material interaction, leading to reduced excessive force, improved accuracy, and higher task success rates~\cite{king2009tactile, pacchierotti2015cutaneous}. Additional studies in manipulation and human–robot interaction report improved precision and more stable contact handling under haptic guidance~\cite{caccavale2023manipulation, bergholz2023benefits}. Modern robotic systems can provide rich sensing cues, including force, slip, and contact geometry, to support these interactions~\cite{li2024comprehensive}. However, effective use of haptic feedback remains limited in practice. Many systems rely on indirect modalities, such as electrotactile or vibrotactile signals, which encode interaction into representations that require interpretation~\cite{lin2026skin, huang2025aerohaptix}. As a result, operators often depend on visual or mediated feedback~\cite{pacchierotti2023cutaneous}, creating a separation between physical interaction and perception, particularly when handling compliant or fragile objects where over-grasping and delayed correction remain common~\cite{Cheng2025AdaptiveGrasping}. These limitations motivate tighter coupling between physical interaction and human perception, providing more direct and interpretable feedback pathways for human-in-the-loop manipulation~\cite{sieber2015multi}.

\begin{figure}[!htbp]
    \centering
    \includegraphics[width=\linewidth]{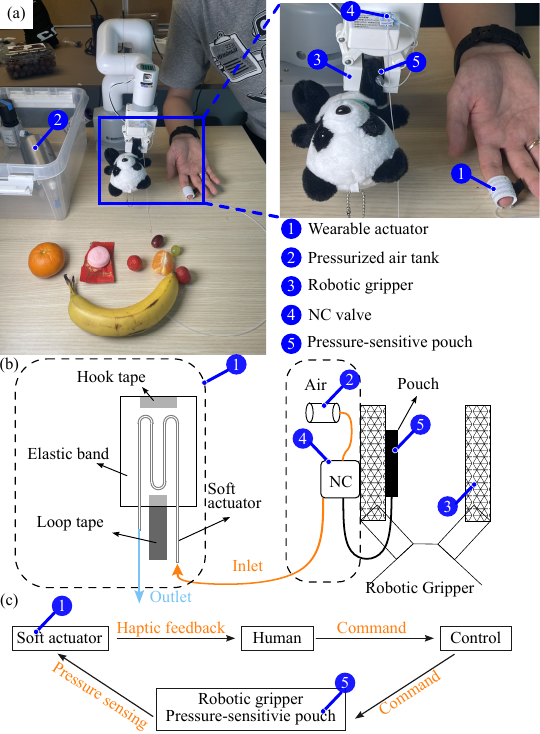}
    \vspace{-2em}
    \caption{System overview of the assistive grasping system. (a) Experimental setup showing the robotic gripper, pressure-sensitive pouch, NC valve, air supply, and wearable pneumatic actuator. (b) Pneumatic interface schematic illustrating airflow paths, valve operation, and mechanical integration. (c) Haptic interaction diagram, where contact-induced pressure changes are conveyed to the user as pneumatic feedback for grasp adjustment.}
    \label{fig:fig1}
\end{figure}

Although haptic sensing is well studied~\cite{okamura2004methods, lepora2026tactile, pacchierotti2023cutaneous}, most approaches rely on electronics and multi-stage pipelines that separate sensing, processing, and feedback, increasing system complexity and introducing intermediate representations. Pneumatic interfaces offer an alternative through fluidic feedback but face limitations in signal fidelity and integration~\cite{king2009tactile}, while recent electronics-free methods focus primarily on sensing rather than direct user interaction~\cite{li2025acoustac}. To address these limitations, we present \textit{TouchDrive}, an electronics-free tactile sensing interface that directly converts contact forces into pneumatic feedback via valve-mediated switching. By embedding sensing and feedback within a physically coupled mechanical loop, the approach removes intermediate representations and provides direct, interpretable cues. Accordingly, our \textbf{contributions} are threefold: (i) a fully passive, electronics-free sensing–actuation system with stable force-to-feedback mapping; (ii) illustrative grasping demonstrations showing gentle manipulation with reduced force on fragile objects; and (iii) validation across heterogeneous robotic platforms with a cost-effective and portable mechanical implementation.

\section{System Design and Fabrication}
\label{sec:Design}

We introduce TouchDrive, a fully passive tactile interface that translates remote contact interactions into user-perceivable pneumatic feedback. The system is designed as a single pneumatic–mechanical loop, in which sensing, signal generation, and tactile feedback are physically coupled. The interface comprises three major components: a pressure-sensitive pouch, a normally closed (NC) valve, and a wearable soft actuator (Fig.~\ref{fig:fig1}(a)(b)). Contact forces are transmitted through the pouch and directly modulate the NC valve, which is co-located with the sensing element on the gripper to minimize response delay caused by grasping. The valve has a compact design (approximately $9\times9\times7.1$ mm$^3$). Valve-mediated switching converts continuous contact forces into pressure signals that are delivered to the user. The system operates under constant pressure and does not require electronic sensing, signal processing, or control. In practice, the interface operates reliably under moderate supply pressures (tested up to approximately 1.5 bar), though its limits are ultimately constrained by the mechanical design of the valve and elastomeric components. The working principle of the haptic feedback loop for assistive grasping is illustrated in Fig.~\ref{fig:fig1}(c), where the human user provides control commands to the robotic gripper during grasping tasks.

\begin{figure}[!htbp]
    \centering
    \includegraphics[width=\linewidth]{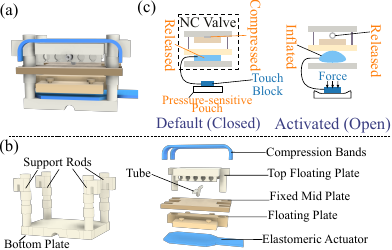}
    \vspace{-2em}
    \caption{Passive NC valve and working principle. (a) Assembled valve module. (b) Exploded view showing individual components and their arrangement. (c) Cross-sectional view illustrating pressure-triggered switching: the soft tube remains closed at rest and opens under sufficient pressure from the sensing pouch, allowing airflow to the wearable actuator.}
    \label{fig:fig2}
\end{figure}

The NC valve (Fig.~\ref{fig:fig2}(a)) comprises a bottom plate, compression bands, floating plates, a silicone supply tube, an elastomeric actuator, and a pressure-sensitive pouch with a pneumatic source (Fig.~\ref{fig:fig2}(b)). It is normally closed by compression bands; when the pouch is compressed, the resulting pressure increase actuates the elastomer, releases the tube, and opens the flow path (Fig.~\ref{fig:fig2}(c)), allowing air to reach the wearable actuator and generate feedback. This passive sensing–actuation loop converts pressure variations into user-perceived feedback, with airflow (typically $\sim$0.1–0.5 SLM) influencing response speed and transmission via silicone tubing producing inflation-based actuation. The system combines off-the-shelf pneumatic components with custom parts, where switching is governed by chamber geometry, elastomer compliance, and preload, yielding a hysteretic response that stabilizes switching. The actuator is fabricated from PDMS (Smooth-Sil 950) with 0.3 mm thickness and active area is $6\times4$ mm$^2$, while the pouch uses heat-shrink tubing sealed with PDMS (Sylgard 184) and silicone adhesive (ELASTOSIL A07). The wearable module is a pneumatic chamber or bent tubing in contact with the finger, where airflow generates tactile cues. The hardware is modular and reconfigurable, using elastic bands for preload tuning and replaceable pouches for different force ranges, with rigid parts, elastomers, and tubing supporting straightforward fabrication; assembly integrates the valve, pouch, and pneumatic lines with preload adjustment, forming a compact, electronics-free design. The pouch is mounted on the gripper tip using adhesive, with a rigid PMMA plate ($4\times4\times2.8$ mm$^2$) to ensure contact, and the NC valve is placed near the contact region to minimize delay. During grasping, contact forces compress the pouch and generate pressure changes that open the valve, producing feedback to the user. The system is deployed on two platforms to evaluate generality across payload scales and hardware configurations: a low-payload manipulator (Elephant Robotics myCobot 280 with an underactuated adaptive gripper, 200 g payload) and an industrial cobot (Dobot Nova 5 with a Robotiq 2F85 gripper, 5 kg payload). For comparison, the Robotiq gripper was evaluated both with its built-in force limit (minimum of approximately 20N) and with our pneumatic interface to assess improvements in controllability and feedback. In each configuration, users regulated grasping through positional adjustments guided by pneumatic feedback. Experiments conducted from predefined poses highlight the portability and simplicity of the interface, demonstrating its effectiveness for assistive grasping regulation.

\section{Experiments Validation}
\label{sec:Experimental}

We evaluate TouchDrive through controlled characterization experiments and real-world grasping tasks. The evaluation targets three aspects: (i) characterization, (ii) effectiveness in assistive grasping, and (iii) effectiveness in assistive grasping compared to a force-limited baseline. 

\subsection{Characterization}
\begin{figure}[!htbp]
    \centering
    \includegraphics{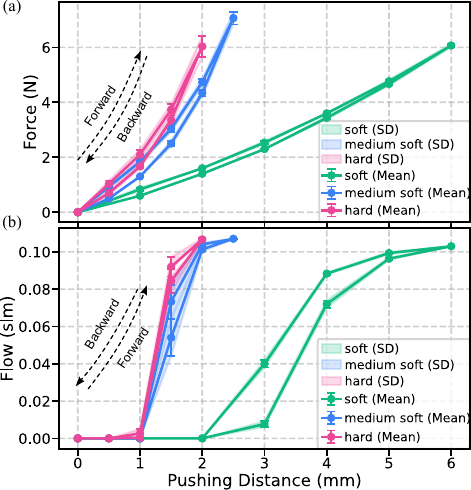}
    \vspace{-1em}
    \caption{Experimental characterization of three silicone cubic with different stiffness (soft, medium soft, hard). (a) Force versus displacement during forwarding and backward, showing steeper responses for harder samples and larger deformation for softer ones. (b) Flow versus displacement, with softer samples requiring greater displacement to initiate flow. Shaded regions and error bars denote standard deviation across trials.}
    \label{fig:fig3}
\end{figure}

Characterization was conducted using a benchtop setup enabling precise, repeatable, and stable control of airflow and pressure under varying loading conditions. The system included a regulated air supply, a precision pressure regulator (VPPI-5L-3-G19-0L12H-V1-S1D, Festo), a flow sensor (SFM5500-0.5SLM, Sensirion), and a force gauge (20N, Sauter) mounted on a linear stage (LSR 330 B, Zaber). A $20\times20\times20$ mm $^3$ cubic sample was placed on a pressure-sensitive pouch with initial contact established prior to actuation. The stage applied cyclic displacements in seven steps of 0.5 mm (hard) or 1 mm (soft and medium soft), while force and flow were recorded at each step to quantify pressure–response behavior (Fig.~\ref{fig:fig3}). Results show material-dependent responses: the soft material exhibited gradual changes in force and flow with displacement, whereas medium-soft and hard materials showed sharper responses, indicating that stiffness strongly affects the rate of force and flow evolution during compression. Furthermore, the observed hysteresis between forward and backward motion confirms that the system behaves as a typical elastomeric structure. The tested materials included soft Ecoflex 00-30 (Smooth-On, $\approx 0.2 \text{MPa}$), medium soft Smooth-sil 950 (Smooth-On, $\approx 2.5 \text{MPa}$), and hard Elastosil RT 675 (Wacker, $\approx 10 \text{MPa}$), with correspondingly increasing tensile strength.

\subsection{Assistive Grasping}
\begin{figure}[!htpb]
    \centering
    \includegraphics[width=0.8\linewidth]{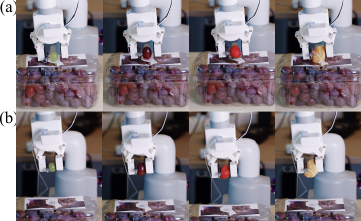}
    \vspace{-1em}
    \caption{Grasping fruits with myCobot using the proposed system. (a) Pre-grasp configuration over mixed fruits. (b) Stable grasping and lifting via pneumatic feedback, demonstrating adaptability across objects.}
    \label{fig:fig4}
\end{figure}

Robot control was implemented in ROS 2 Humble~\cite{macenski2022robot}, handling user inputs (joystick or keyboard), joint-state feedback, and trajectory execution, while motion planning was performed in MoveIt 2~\cite{coleman2014reducing} using Cartesian paths from a home pose to predefined grasp poses with the tool frame as the end effector; planned trajectories were published and converted into joint commands, with gripper opening logged via ROS topics. The myCobot 280 equipped with an adaptive two-finger gripper was used for fruit grasping, where pneumatic feedback guided the user’s decision to continue or stop closing the gripper. Participants wore an eye mask to remove visual cues and incrementally controlled the position-driven gripper based solely on tactile feedback. The system was evaluated on fruits with varying geometry and compliance, including grapes, clementine segments, and strawberries, chosen for their softness and susceptibility to damage. Objects were randomly placed on a transparent box to vary pose and contact conditions (Fig.~\ref{fig:fig4}). After the user indicated a secure grasp, the robot lifted the object to assess grasp stability.

\begin{figure}[!htpb]
    \centering
    \includegraphics[width=0.8\linewidth]{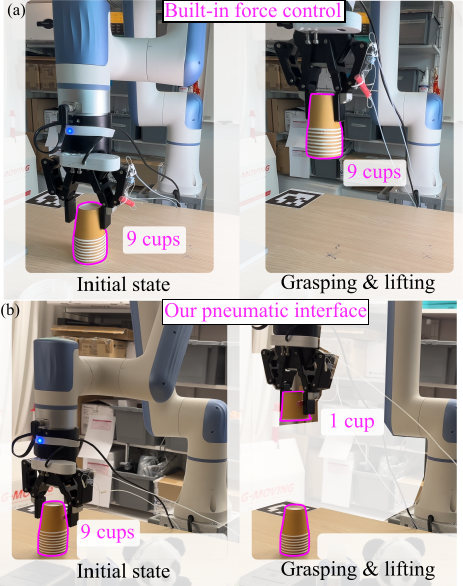}
    \vspace{-1em}
    \caption{Grasping a stack of nine paper cups; nine cups correspond to the approximate upper limit imposed by inter-cup friction. (a) Using the gripper’s built-in force control (minimum 20 N), resulting in over-compression and lifting of all cups. (b) Using the pneumatic sensing interface, enabling force regulation and selective grasping, resulting in a selective single-cup grasp at sub-newton contact scale.}
    \label{fig:fig5}
\end{figure}

As shown in (Table~\ref{tab:successrate}), the Dobot Nova 5 equipped with the Robotiq 2F85 gripper was evaluated across 16 everyday objects spanning rigid, deformable, and fragile categories, with the pneumatic interface providing tactile feedback to guide grasp termination. The binary-plus-partial scheme, where 1 denotes a stable grasp, 0.5 a grasp lost during transport, and 0 a grasp failure, captures both contact success and functional reliability, providing a more informative metric than binary success alone. The system achieved a mean score of 0.94 across all Dobot trials, with 12 out of 16 items reaching a perfect score of 1.0, demonstrating consistent and reliable performance. Notably, delicate items such as eggs, small tomatoes, grapes, and strawberries were grasped successfully in all five trials each, highlighting the sensitivity of the passive pneumatic feedback in preventing over-grasping. The paper-cup task resulted in an average of 1.6 cups held per trial (Fig.~\ref{fig:fig5}(b)), demonstrating that users could modulate grip force through feedback alone to stack or retain multiple cups; for reference, a stack of 10 cups weighs roughly 10 g, and nine cups represent approximately the upper limit imposed by inter-cup friction. In comparison, the gripper operating with its built-in force control (minimum 20N) was able to lift all nine cups in the stack (Fig.~\ref{fig:fig5}(a)), indicating that the original gripper lacks the ability to apply as gentle and finely tuned a grasp as our interface enables. Lower scores were observed for the screwdriver (0.5), the measuring tape (0.6), and plastic box (0.6). An important point to note is that users often interpret the initial tactile feedback, when the gripper first makes contact with the object, as a signal that the grasp is sufficient; however, in many cases, a firmer grasp is required to securely hold the object rather than lifting upon contact, especially for heavier or slippery items. One potential factor to the lower scores is that the interface is highly sensitive, so once contact occurs, users may lift prematurely.

In contrast, the myCobot platform, tested on four fruit items using the underactuated gripper, achieved a perfect score of 1.0 across all 20 trials. This result demonstrates that the interface is portable and performs reliably across hardware systems with significantly different payload capacities and stiffness characteristics. 
The overall results confirm that the proposed passive pneumatic sensing–actuation loop enables effective assitive grasping regulation without electronics, demonstrating broad applicability across object types, robot platforms, and user-applied force ranges.

\begin{table}[!htbp]
\centering
\caption{Grasping items and success rate trials.}
\vspace{-1em}
\begin{tabular}{c|cccc}
\hline
ID & Item & Robot & Test & Score \\
\hline
1&Eraser  & (Dobot)  & 1;1;1;1;1; & 1  \\
2&Measuring tape  & (Dobot) & 1;1;0;1;0; & 0.6  \\
3&Aluminum profile & (Dobot) & 1;1;1;1;1; & 1  \\
4&Plastic box  & (Dobot) & 0;1;1;0;1; & 0.6  \\
5&Marker pen  & (Dobot)& 1;1;1;1;1; & 1 \\
6&Hand cream  & (Dobot) & 1;1;1;1;1; & 1  \\
7&3D printed octopus  & (Dobot) & 1;1;1;1;1; & 1  \\
8&Pandas toy  & (Dobot) & 1;1;1;0.5;1; & 0.9  \\
9&Paper cups  & (Dobot)& 1;2;1;2;2;& 1.6 \\
10&Screwdriver  & (Dobot) & 0.5;1;0.5;0;0.5; & 0.5 \\
11&Plastic beaker  & (Dobot) & 1;1;1;1;1; & 1 \\
12&Plastic bottle  & (Dobot) & 1;1;1;1;0.5; & 0.9 \\
13&Small tomato  & (Dobot) & 1;1;1;1;1; & 1 \\
14&Grape  & (Dobot)&  1;1;1;1;1;& 1 \\
15&Stawberry  & (Dobot)& 1;1;1;1;1;&1 \\
16&Egg  & (Dobot)& 1;1;1;1;1; & 1 \\
17&Green grape  & (myCobot) & 1;1;1;1;1; & 1 \\
18&Purple grape &(myCobot)  & 1;1;1;1;1; & 1 \\
19&Strawberry & (myCobot) &  1;1;1;1;1;& 1 \\
20&Clementine & (myCobot) & 1;1;1;1;1; & 1 \\
\hline
\end{tabular}
\label{tab:successrate}
\end{table}

\section{Conclusion}
\label{sec:Discussion}
This work presented TouchDrive, an electronics-free tactile sensing interface for assistive grasping that directly converts contact forces into user-perceivable pneumatic feedback through valve-mediated switching. By embedding sensing, signal generation, and feedback within a passive mechanical loop, the interface provides stable and repeatable tactile cues without electronics or processing. Through grasping demonstrations, we showed that the pneumatic interface enables users to regulate contact forces and achieve gentle manipulation of compliant and fragile objects. The interface was validated across heterogeneous robotic platforms, including a low-payload manipulator and an industrial cobot, demonstrating its versatility and scalability. With minimal hardware requirements and no reliance on electronic components, TouchDrive offers a cost-effective, portable, and mechanically simple solution for assistive grasping.

\bibliographystyle{IEEEtran}
\bibliography{IEEEabrv,reference}

\end{document}